# Semi-Parallel Deep Neural Network (SPDNN) Hybrid Architecture, First Application on Depth from Monocular Camera

Shabab Bazrafkan,[a, ¶] Hossein Javidnia,[a,*, ¶] Joseph Lemley,[a] Peter Corcoran[a]

[a]National University of Ireland Galway, College of Engineering, Department of Electronic Engineering, University Road, Galway, Ireland

**Abstract**. Deep neural networks are applied to a wide range of problems in recent years. In this work, Convolutional Neural Network (CNN) is applied to the problem of determining the depth from a single camera image (monocular depth). Eight different networks are designed to perform depth estimation, each of them suitable for a feature level. Networks with different pooling sizes determine different feature levels. After designing a set of networks, these models may be combined into a single network topology using graph optimization techniques. This "Semi Parallel Deep Neural Network (SPDNN)" eliminates duplicated common network layers, and can be further optimized by retraining to achieve an improved model compared to the individual topologies. In this study, four SPDNN models are trained and have been evaluated at 2 stages on the KITTI dataset. The ground truth images in the first part of the experiment are provided by the benchmark, and for the second part, the ground truth images are the depth map results from applying a state-of-the-art stereo matching method. The results of this evaluation demonstrate that using post-processing techniques to refine the target of the network increases the accuracy of depth estimation on individual mono images. The second evaluation shows that using segmentation data alongside the original data as the input can improve the depth estimation results to a point where performance is comparable with stereo depth estimation. The computational time is also discussed in this study.

**\***Corresponding Author**,** E-mail: h.javidnia1@nuigalway.ie
¶These authors contributed equally to this work.

## 1 Introduction

Computing pixel depth values provides a basis for understanding the 3D geometrical structure of images. As it has been presented in recent research [1], using stereo images provides an accurate depth due to the advantage of having local correspondences; however, the processing times of these methods is still an open issue.

To solve this problem, it has been suggested to use single images to compute the depth values, but extracting depth from monocular images requires extracting a large number of cues from the global and local information in the image. Using a single camera is more convenient in industrial applications. Stereo cameras require detailed calibration and many industrial use cases already employ single cameras – e.g. security monitoring, automotive & consumer vision systems, and camera infrastructure for traffic and pedestrian management in smart cities. These and other smart-vision applications can greatly benefit from accurate monocular depth analysis. This challenge has been studied for a decade and is still an open research problem.

Recently the idea of using neural networks to solve this problem has attracted attention. In this paper, we tackle this problem by employing a Deep Neural Network (DNN) equipped with semantic pixel-wise segmentation utilizing our recently published disparity post-processing method.

This paper also introduces the use of *Semi Parallel Deep Neural Networks* (SPDNN). A SPDNN is a semi-parallel network topology developed using a graph theory optimization of a set of independently optimized CNNs, each targeted at a specific aspect of the more general classification problem. In [2] [3] the effect of SPDNN approach on increasing convergence and improving model generalization is discussed. For the depth from monocular vision problem a fully-connected topology, optimized for fine features, is combined with a series of max-pooled topologies (2×2, 4×4 and 8×8) each optimised for coarser image features. The optimized SPDNN topology is re-trained on the full training dataset and converges to an improved set of network weights.

It is worth mentioning that this network design strategy is not limited to the 'depth from monocular vision' problem, and further application examples and refinements will be developed in a series of future publications, currently in press.

*1.1 Depth Map*

Deriving the 3D structure of an object from a set of 2D points is a fundamental problem in computer vision. Most of these conversions from 2D to 3D space are based on the depth values computed for each 2D point. In a depth map, each pixel is defined not by color, but by the distance between an object and the camera. In general, depth computation methods are divided into two categories:

1- Active methods
2- Passive methods

Active methods involve computing the depth in the scene by interacting with the objects and the environment. There are different types of active methods, such as light-based depth estimation, which uses the active light illumination to estimate the distance to different objects [4]. Ultrasound and time-of-flight (ToF) are other examples of active methods. These methods use the known speed of the wave to measure the time an emitted pulse takes to arrive at an image sensor [5].

Passive methods utilize the optical features of captured images. These methods involve extracting the depth information by computational image processing. In the category of passive methods, there are two primary approaches a) Multi-view depth estimation, such as depth from stereo, and b) Monocular depth estimation.

*1.2 Stereo Vision Depth*

Stereo matching algorithms can be used to compute depth information from multiple images. By using the calibration information of the cameras, the depth images can be generated. This depth information provides useful data to identify and detect objects in the scene [6].

In recent years, many applications, including time-of-flight [7,8], structured light [9], and Kinect were introduced to calculate depth from stereo images. Stereo vision algorithms are generally divided into two categories: Local and Global. Local algorithms were introduced as statistical methods that use the local information around a pixel to determine the depth value of the given pixel. These kinds of methods can be used for real-time applications if they are implemented efficiently. Global algorithms try to optimize an energy function to satisfy the depth estimation problem through various optimization techniques [10].

In terms of computation, global methods are more complex than local methods, and they are usually impractical for real-time applications. Despite these drawbacks, they have the advantage in being more accurate than local methods. This advantage recently attracted considerable attention in the academic literature [11,12].

For example, the global stereo model proposed in [11] works by converting the image into a set of 2D triangles with adjacent vertices. Later, the 2D vertices are converted to a 3D mesh by computing the disparity values. To solve the problem of depth discontinuities, a two-layer Markov Random Field (MRF) is employed. The layers are fused with an energy function allowing the method to handle the depth discontinuities. The method has been evaluated on the new Middlebury 3.0 benchmark [12] and it was ranked the most accurate at the time of the paper's publication based on the average weight on the bad 2.0 index.

Another global stereo matching algorithm, proposed in [13], makes use of the texture and edge information of the image. The problem of large disparity differences in small patches of non-textured regions is addressed by utilizing the color intensity. In addition, the main matching cost function produced by a CNN is augmented using the same color-based cost. The final results are post-processed using a 5×5 median filter and a bilateral filter. This adaptive smoothness filtering technique is the primary reason for the algorithm's excellent performance and placement in the top of the Middlebury 3.0 benchmark [12].

Many other methods have been proposed for stereo depth, such as PMSC [12], GCSVR [12], INTS [14], MDP [15], ICSG [16], which all aimed to improve the accuracy of the depth estimated from stereo vision, or to introduce a new method to estimate the depth from a stereo pair. However, there is always a trade-off between accuracy and speed for stereo vision algorithms.

Table 1 Comparison of the performance time between the most accurate stereo matching algorithms

| Algorithm | Time/MP (s) | W × H (ndisp) | Programming Platform | Hardware |
|---|---|---|---|---|
| PMSC [12] | 453 | 1500 × 1000 (<= 400) | C++ | i7-6700K, 4GHz-GTX TITAN X |
| MeshStereoExt [11] | 121 | 1500 × 1000 (<= 400) | C, C++ | 8 Cores-NVIDIA TITAN X |
| APAP-Stereo [12] | 97.2 | 1500 × 1000 (<= 400) | Matlab+Mex | i7 Core 3.5GHz, 4 Cores |
| NTDE [13] | 114 | 1500 × 1000 (<= 400) | n/a | i7 Core, 2.2 GHz-Geforce GTX TITAN X |
| MC-CNN-acrt [17] | 112 | 1500 × 1000 (<= 400) | n/a | NVIDIA GTX TITAN Black |
| MC-CNN+RBS [18] | 140 | 1500 × 1000 (<= 400) | C++ | Intel(R) Xeon(R) CPU E5-1650 0, 3.20GHz, 6 Cores- 32 GB RAM-NVIDIA GTX TITAN X |
| SNP-RSM [12] | 258 | 1500 × 1000 (<= 400) | Matlab | i5, 4590 CPU, 3.3 GHz |
| MCCNN_Layout [12] | 262 | 1500 × 1000 (<= 400) | Matlab | i7 Core, 3.5GHz |
| MC-CNN-fst [17] | 1.26 | 1500 × 1000 (<= 400) | n/a | NVIDIA GTX TITAN X |
| LPU [12] | 3523 | 1500 × 1000 (<= 400) | Matlab | Core i5, 4 Cores- 2xGTX 970 |
| MDP [15] | 58.5 | 1500 × 1000 (<= 400) | n/a | 4 i7 Cores, 3.4 GHz |
| MeshStereo [11] | 54 | 1500 × 1000 (<= 400) | C++ | i7-2600, 3.40GHz, 8 Cores |
| SOU4P-net [12] | 678 | 1500 × 1000 (<= 400) | n/a | i7 Core, 3.2GHz-GTX 980 |
| INTS [14] | 127 | 1500 × 1000 (<= 400) | C, C++ | i7 Core, 3.2 GHz |
| GCSVR [12] | 4731 | 1500 × 1000 (<= 400) | C++ | i7 Core, 2.8GHz-Nvidia GTX 660Ti |
| JMR [12] | 11.1 | 1500 × 1000 (<= 400) | C++ | Core i7, 3.6 GHz-GTX 980 |
| LCU [12] | 9572 | 750 × 500 (<= 200) | Matlab, C++ | 1 Core Xeon CPU, E5-2690, 3.00 GHz |
| TMAP [19] | 1796 | 1500 × 1000 (<= 400) | Matlab | i7 Core, 2.7GHz |
| SPS [12] | 49.4 | 3000 × 2000 (<= 800) | C, C++ | 1 i7 Core, 2.8GHz |
| IDR [20] | 0.36 | 1500 × 1000 (<= 400) | CUDA C++ | NVIDIA GeForce TITAN Black |

Table 1 shows an overview of the average normalized time by the number of pixels (sec/megapixels) of the most accurate stereo matching algorithms as they are ranked by the Middlebury 3.0 benchmark, based on the "bad 2.0" metric. The ranking is on the test dense set. This comparison illustrates that obtaining an accurate depth from a stereo pair requires significant processing power. These results demonstrate that today, these methods are too resource intensive for real-time applications like street sensing or autonomous navigation due to their demand for processing resources.

To decrease the processing power of stereo matching algorithms, researchers recently began to work on depth from monocular images. Such algorithms estimate depth from a single camera while keeping the processing power low.

*1.3 Deep Learning*

DNN (Deep Neural Networks) are among the most recent approaches in pattern recognition science that are able to handle highly non-linear problems in classification and regression. These models use consecutive non-linear signal processing units in order to mix and re-orient their input data to give the most representative results. The DNN structure learns from the input and then it generalizes what it learns into data samples it has never seen before [21]. The typical deep neural network model is composed of one or more convolutional, pooling, and fully connected layers accompanied by different regularization tasks. Each of these units is as follows:

**Convolutional Layer**: This layer typically convolves the 3D image $I$ with the 4D kernel $W$ and adds a 3D bias term $b$ to it. The output is given by:

$$P = I * W + b \qquad (1)$$

where $*$ operator is nD convolution and $P$ is the output of the convolution. During the training process, the kernel and bias parameters are updated in a way that optimizes the error function of the network output.

**Pooling Layer**: The pooling layer applies a (usually) non-linear transform (Note that the average pooling is a linear transform, but the more popular max-pooling operation is non-linear) on the input image which reduces the spatial size of the data representation after the operation.

It is common to put a pooling layer after each convolutional layer. Reducing the spatial size leads to less computational load and also prevents over-fitting. The reduced spatial size also provides a certain amount of translation invariance.

**Fully Connected Layer**: Fully connected layers are the same as classical Neural Network (NN) layers, where all the neurons in a layer are connected to all the neurons in their subsequent layer. The neurons give the summation of their input, multiplied by their weights, passed through their activation functions.

**Regularization**: Regularization is often used to prevent overfitting of a neural network. One can train a more complex network (more parameters) with regularization and prevent over-fitting. Different kinds of regularization methods have been proposed. The most important ones are weight regularization, drop-out [22], and batch normalization [23]. Each regularization technique is suitable for specific applications, and no single technique works for every task.

*1.4 Monocular Vision Depth*

Depth estimation from a single image is a fundamental problem in computer vision and has potential applications in robotics, scene understanding, and 3D reconstruction. This problem remains challenging because there are no reliable cues for inferring depth from a single image. For example, temporal information and stereo correspondences are missing from such images.

As the result of the recent research, deep Convolutional Neural Networks (CNN) are setting new records for various vision applications. A deep convolutional neural field model for estimating depths from a single image has been presented in [24] by reformulating the depth estimation into a continuous conditional random field (CRF) learning problem. The CNN employed in this research was composed of 5 convolutional and 4 fully-connected layers. At the first stage of the algorithm, the input image was over-segmented into superpixels. The cropped image patch centered on its centroid was used as input to the CNN. For a pair of neighboring superpixels, a number of similarities were considered and were used as the input to the fully connected layer. The output of these 2 parts was then used as input to the CRF loss layer. As a result, the time required for estimating the depth from a single image using the trained model decreased to 1.1 seconds on a desktop PC equipped with NVIDIA GTX 780 GPU with 6GB memory.

It has been found that the superpixelling technique of [24] is not a good choice to initialize the disparity estimation from mono images because of the lack of the monocular visual cues such as texture variations and gradients, defocus or color/haze in some parts of the image. To solve this issue an MRF learning algorithm has been implemented to capture some of these monocular cues [25]. The captured cues were integrated with a stereo system to obtain better depth estimation than the stereo system alone. This method uses a fusion of stereo + mono depth estimation.

At small distances, the algorithm relies more on stereo vision, which is more accurate than monocular vision. However, at further distances, the performance of stereo degrades; and the algorithm relies more on monocular vision.

The problem of depth estimation from monocular images has been also studied in [26] where a network is designed with two components. First, the global structure of the scene is estimated and later refined using local information. Although this approach enables the early idea of estimating monocular depth using CNNs, the output depth maps do not clearly represent the geometrical structure of the scene.

In another approach [27], an unsupervised convolutional encoder is trained to estimate the depth from monocular images. The depth is estimated considering the small motion between two images (stereo set as input and target). Later, the inverse warp of the target image is generated using the predicted depth and the known displacement between cameras which results in reconstructing the source image. In a similar research [28], an unsupervised CNN is trained by exploiting Epipolar geometry constraints to estimate disparity from single images. The idea is to learn a function that is able to reconstruct one image from the other, by utilizing a calibrated pair of binocular cameras. A left-right disparity consistency loss is also introduced which combines smoothness, reconstruction, and left-right disparity consistency terms and keeps the consistency between the disparities produced relative to both the left and right images.

*1.5 Paper Overview*

In this paper, a DNN is presented to estimate depth from monocular cameras. The depth map from the stereo sets are estimated using the same approach as [29] and they are used as the target to train the network while using information from a single image (the left image in the stereo set) as input. Four models are trained and evaluated to estimate the depth from single camera images. The network structure for all the models is same. In the first case, the input is simply the original image. In the second case, the first channel is the original image and the second channel is its segmentation map. For each of these two cases, one of two different targets are used; specifically, these targets were the stereo depth maps with or without post-processing explained in [29]. Fig. 1 shows the overview of the general approach used in this paper.

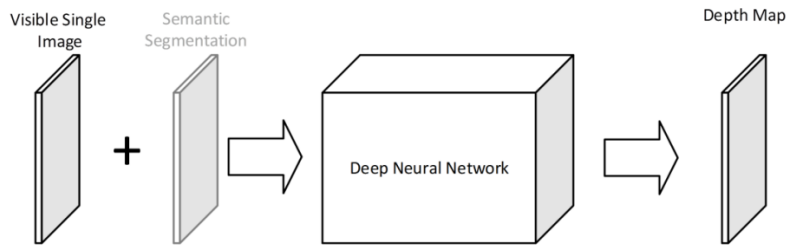

**Fig. 1** The overview of the trained models in this paper. The semantic segmentation is just used in two experiments

*1.6 Contributions*

In this paper two major contributions are presented:

1- A method to mix and merge several deep neural networks called "Semi Parallel Deep Neural Network (SPDNN)", described in detail in Appendix A.
2- The application of deep neural networks and SPDNN on estimating depth from a monocular camera.

The rest of the paper is organized as follows: In the next section the network structure, database preparation, and the training process are presented. Sec 3 discusses the results and evaluation of the proposed method. The conclusion and discussions are presented in the last section.

## 2 Methodology

*2.1 Network Structure*

*2.1.1 Semi-Parallel Deep Neural Network (SPDNN)*

This paper introduces the SPDNN concept, inspired by graph optimization techniques. In this method, several deep neural networks are parallelized and merged in a novel way that facilitates the advantages of each. The final model is trained for the problem. [2] [3] show that using this method increases the convergence and generalization of the model compared to alternatives.

The merging of multiple networks using SPDNN is described in the context of the current depth mapping problem. In this particular problem, eight different networks were designed for the depth estimation task. These are described in detail in Appendix A. None of these networks on their own gave useful results on the depth analysis problem. However, it was noticed that each network tended to perform well on certain aspects of this task while failing at others. This led to the idea that it would be advantageous to combine multiple individual networks and train them in a parallelized architecture. Our experiments showed that better output could be achieved by merging the networks and then training them concurrently.

*2.1.1.1   The Combined Model/Architecture*

The process of the network design is discussed in detail in Appendix A. In the final model presented in Fig. 2, the input image is first processed in four, parallel fully convolutional sub-networks with different pooling sizes. This provides the advantages of different networks with different pooling sizes at the same time. The outputs of these four sub-networks are concatenated in two different forms; one to pool the larger images to be the same size as the smallest image in the previous part, and the other one is to un-pool the smaller images of the previous part to be the same size as the largest image.

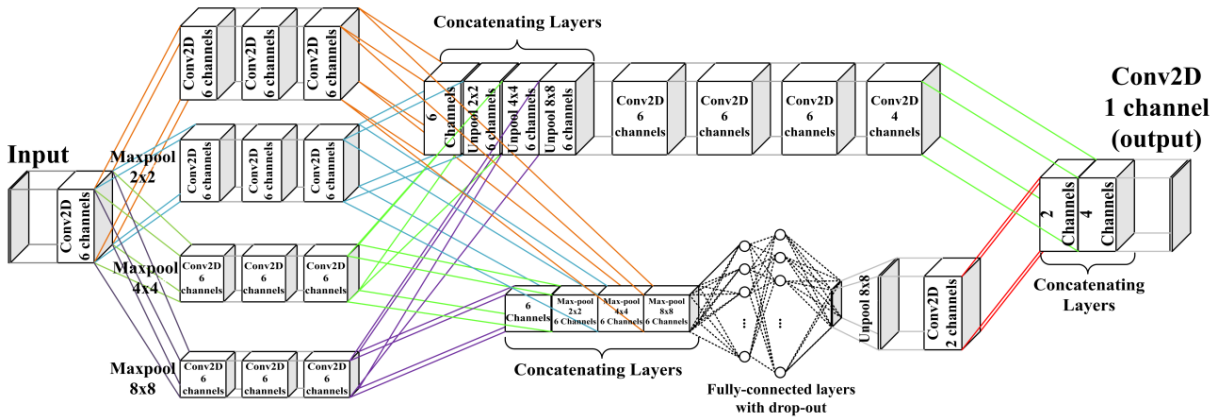

**Fig. 2** The model designed for the depth estimation from monocular images.

After merging these outputs, the data is led to 2 different networks. One is the fully convolutional network to deepen the learning and release more abstract features of the input, and the other network is an auto-encoder network with different architecture for encoder and decoder.

It is mentioned in the network design section in Appendix A that, having a fully connected layer in the network is crucial for the reasonable estimation of the image's depth which is provided in the bottleneck of the autoencoder. The results from the autoencoder and the fully convolutional sub-network are again merged in order to give a single output after applying a one channel convolutional layer.

In order to regularize the network, prevent overfitting and increase the convergence, batch normalization [23] is applied after every convolutional layer, and the drop-out technique [22] is used in fully connected layers. The experiments in this paper show that using weight regularization in the fully connected layers gives slower convergence; therefore, this regularization was eliminated from the final design. All the nonlinearities in the network are the ReLU nonlinearity, which is widely used in deep neural networks, except the output layer, which took advantage of the sigmoid nonlinearity. The value repeating technique was used in the un-pooling layer due to non-specificity of the corresponding pooled layer in the decoder part of the auto-encoder sub-network.

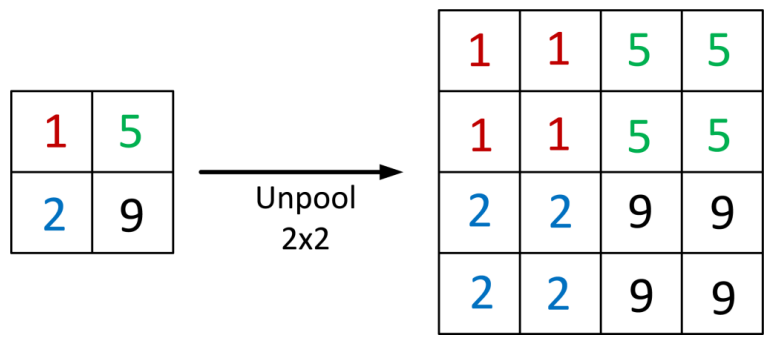

**Fig. 3** The repeating technique used in un-pooling layers.

The value repeating technique, illustrated in Fig. 3, involves repeating the value from the previous layer in order to obtain the un-pooled image. The figure shows the 2×2 un-pooling, and the process is the same for other un-pooling sizes.

*2.2 Database*

In this paper, the KITTI Stereo 2012, 2015 datasets [30] are used for training and evaluation of the network. The database is augmented by vertical and horizontal flipping to expand the total size to 33,096 images. 70% of this dataset is used for training, 20% for validation and 10% for testing. Each model is trained for two sets of input samples and two sets of output targets. The input and target preparation are explained in the following sections.

*2.2.1 Data Preparation*

*2.2.1.1 Input Preparation*

Two different sets have been used as the input of the network. The first set includes the visible images given by the left camera. The second set is the visible image + the semantic segmentation of the corresponding input. This gives the opportunity of investigating the segmentation influence on the depth estimation problem. The segmentation map for each image is calculated by employing the well-known model "SegNet" [31,32]. This model is one of the most successful recent implementations of DNN for semantic pixel-wise image segmentation and has surpassed other configurations of Fully Convolutional Networks (FCN) both in accuracy and simplicity of implementation. A short description on SegNet is given in Appendix B.

In our experiments, SegNet was trained using Stochastic Gradient Descent (SGD) with learning rate 0.1 and momentum 0.9. In this paper, the Caffe implementation of SegNet has been employed for training purposes [33]. The gray-scale CamVid road scene database (360×480) [34] has been used in the training step.

*2.2.1.2  Target Preparation*

The targets for training the network are generated from the stereo information using the Adaptive Random Walk with Restart algorithm [35]. The output of the stereo matching algorithm suffers from several artifacts which are addressed and solved by a post-processing method in [29]. In the present experiments, both depth maps (before post-processing and after post-processing) are used independently as targets. The post-processing procedure is based on the mutual information of the RGB image (used as a reference image) and the initial estimated depth image. This approach has been used to increase the accuracy of the depth estimation in stereo vision by preserving the edges and corners in the depth map and filling in the missing parts. The method was compared with the top 8 depth estimation methods in the Middlebury benchmark [12] at the time the paper was authored. Seven metrics, including Mean Square Error (MSE), Root Mean Square Error (RMSE), Peak Signal-to-Noise Ratio (PSNR), Signal-to-Noise Ratio (SNR), Mean Absolute Error (MAE), Structural Similarity Index (SSIM) and Structural Dissimilarity Index (DSSIM) were used to evaluate the performance of each method. The evaluation ranked the method as $1^{st}$ in 5 metrics and $2^{nd}$ and $3^{rd}$ in other metrics

*2.3  Training*

As described in Sec 2.2.1.1 and 2.2.1.2 there are two separate sets as input and two separate sets as targets for the training process. This will give four experiments in total as follows:

1- **Experiment 1**: Input: Left Visible Image + Pixel-wise Segmented Image.  Target: Post-Processed Depth map

2- **Experiment 2**: Input: Left Visible Image. Target: Post-Processed Depth map.

3- **Experiment 3**: Input: Left Visible Image + Pixel-wise Segmented Image. Target: Depth map.

4- **Experiment 4**: Input: Left Visible Image. Target: Depth map.

The images are resized to 80×264 pixels during the whole process. Training is done on a standard desktop with an NVIDIA GTX 1080 GPU with 8GB memory.

In the presented experiments, the mean square error value between the output of the network and the target values have been used as the loss function, and the Nestrov momentum technique [36] with learning rate 0.01 and momentum 0.9 has been used to train the network. The Training and Validation Loss for each of these experiments are shown in Fig. 4 and Fig. 5 respectively.

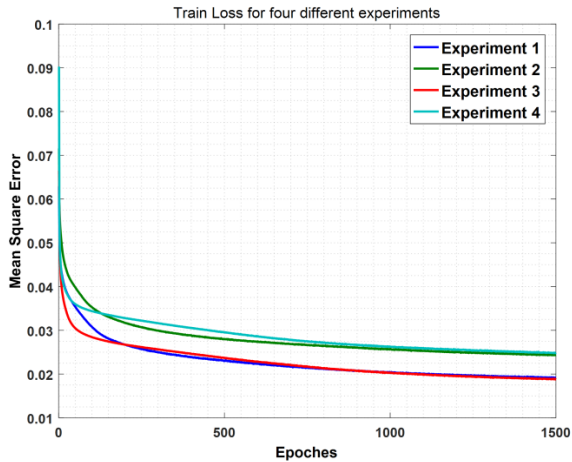
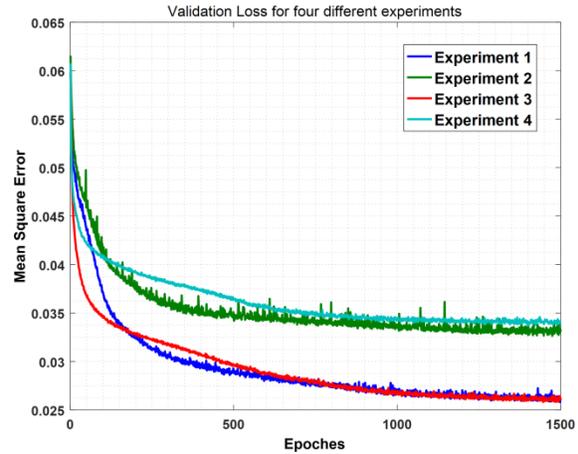

**Fig. 4** Train loss for each experiment

**Fig. 5** Validation loss for each experiment

These figures show that using the Post-Processed Depth map as the target results in lower loss values, which means that the network was able to learn better features in those experiments, while semantic segmentation decreases the error only marginally

# 3 Results and Evaluations

The evaluation in this paper has been done in 4 parts. In the first two parts, the four experiments given in Sec 2.3 are compared to each other, given different ground truths. The third part compares the proposed method to a stereo matching method and the last part shows the comparison against the state of the art monocular depth estimation method. For evaluation purposes, 8 metrics including PSNR, MSE (between 0 and 1), RMSE (between 0 and 1), SNR, MAE (between 0 and 1), Structural Similarity Index (SSIM)(between 0 and 1) [37], Universal Quality Index (UQI) (between 0 and 1) [38] and Pearson Correlation Coefficient (PCC) (between -1 and 1) [39] are used. For the metrics PSNR, SNR, SSIM, UQI, and PCC the larger value indicates better performance, and for MSE, RMSE, and MAE, the lower value indicates better performance. PSNR, MSE, RMSE, MAE, and SNR represent the general similarities between two objects. UQI and SSIM are structural similarity indicators and PCC represents the correlation between two samples. To the best of our knowledge, there have been no other attempts at estimating depth from a mono camera on the KITTI benchmark.

## 3.1 Comparing Experiments Given Benchmark Ground Truth

The KITTI database came with a depth map ground truth generated by a LIDAR scanner.

**Table 2** Numerical comparison of the models given the benchmark's ground truth

|      | Exp. 1  | Exp. 2  | Exp. 3  | Exp. 4  |
|------|---------|---------|---------|---------|
| PSNR | **14.3424** | 13.7677 | 13.8333 | 13.8179 |
| MSE  | **0.0382** | 0.0436 | 0.0435 | 0.0439 |
| RMSE | **0.1937** | 0.2069 | 0.206  | 0.2066 |
| SNR  | 4.4026  | 3.8279  | **6.1952** | 6.1798 |
| MAE  | **0.1107** | 0.1212 | 0.1236 | 0.1234 |
| SSIM | **0.9959** | 0.9955 | 0.9955 | 0.9955 |
| UQI  | 0.9234  | **0.9252** | 0.9053 | 0.9064 |
| PCC  | 0.7687  | **0.8485** | 0.7702 | 0.7729 |

The test set has been forward propagated through the four different models trained in the four experiments, and the output of the networks has been compared to the benchmark ground truth. The results are shown in Table 2. The best value for each metric is presented in bold.

Figs. 6-8 represent the color-coded depth maps computed by the trained models using the proposed DNN, where the dark red and dark blue parts represent closest and furthest points to the camera respectively. On the top right of each figure, the ground truth given by the benchmark is illustrated. For visualization purposes, all of the images presented in this section are upsampled using Joint Bilateral Upsampling [40]. The results show that using semantic segmentation along with the visible image as input will improve the model marginally. Using the post-processed target in the training stage helps the model to converge to more realistic results.

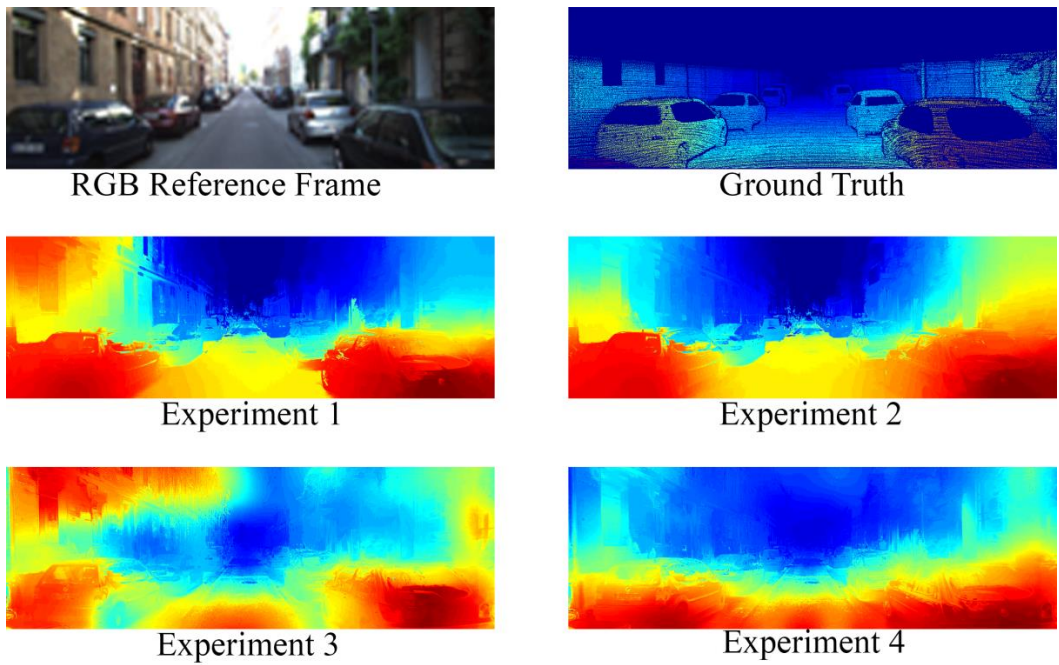

**Fig. 6** Estimated depth maps from the trained models – example 1

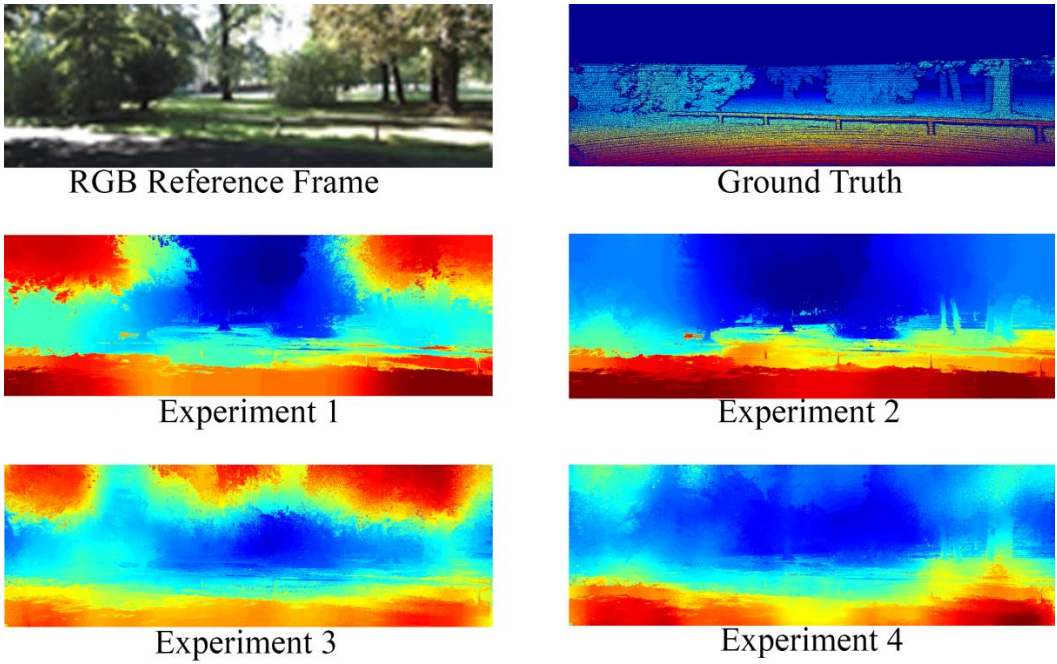

**Fig. 7** Estimated depth maps from the trained models – example 2

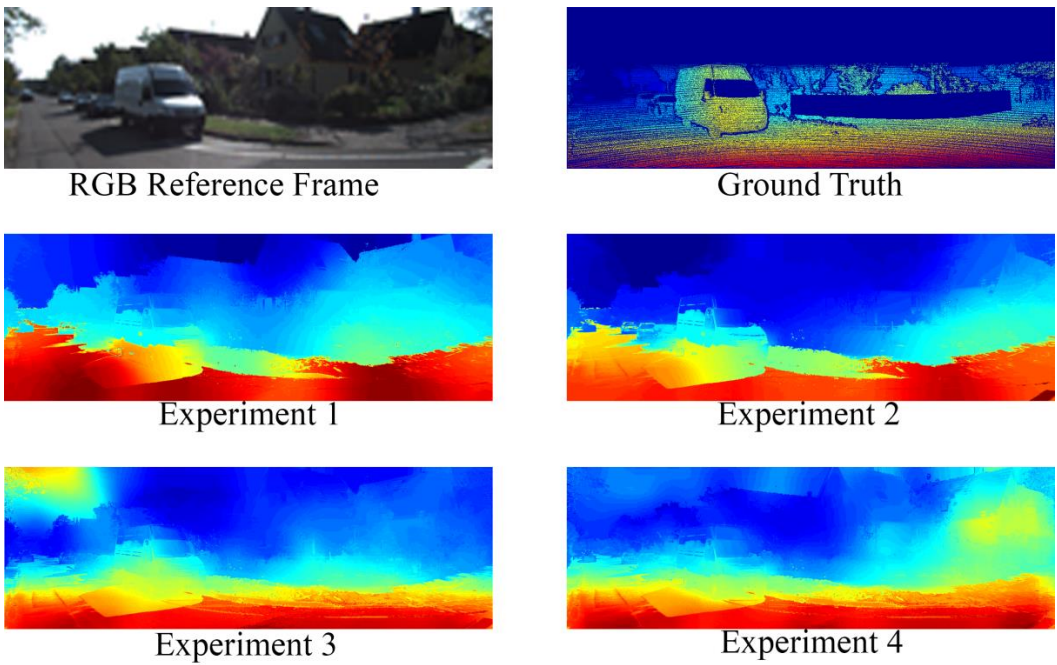

**Fig. 8** Estimated depth maps from the trained models – example 3

As it is illustrated in Figs. 6–8, the depth map generated in experiment 1 contains more structural details, and more precise, less faulty depth levels compared with the other experiments.

In general, the presented models in this paper are able to handle occlusions and discontinuities at different depth levels.

*3.2 Comparing Experiments Given the Ground Truth from Stereo Matching*

In this section, proposed models are compared to see which one produces closer results to the target value. This gives an idea whether using deep learning techniques on the mono camera can produce reasonable results or not.

**Table 3** Numerical comparison of the models given the ground truth from stereo matching

|      | Exp. 1  | Exp. 2  | Exp. 3  | Exp. 4  |
|------|---------|---------|---------|---------|
| PSNR | **15.0418** | 14.1895 | 13.3819 | 14.0491 |
| MSE  | **0.0378**  | 0.0447  | 0.0535  | 0.0441  |
| RMSE | **0.1854**  | 0.203   | 0.2223  | 0.2039  |
| SNR  | **8.822**   | 7.9696  | 5.4271  | 6.0943  |
| MAE  | **0.1442**  | 0.1581  | 0.1673  | 0.153   |
| SSIM | **0.9952**  | 0.9943  | 0.994   | 0.9951  |
| UQI  | **0.8401**  | 0.8369  | 0.7951  | 0.8178  |
| PCC  | **0.8082**  | 0.795   | 0.704   | 0.6919  |

Images in the test set have been forward propagated through the models trained in Sec 2.3, and the outputs are compared with the depth map generated by [29]. The numerical results are shown in Table 3.

The best value for each metric is presented in bold. Figs. 9-11 represent the color-coded depth maps computed by the trained models using the proposed DNN, where the dark red and dark blue parts represent closest and furthest points to the camera respectively. On the top right of each figure, the ground truth calculated by [29] is illustrated. For visualization purposes, all of the images presented in this section are upsampled using Joint Bilateral Upsampling [40]. The results show that using semantic segmentation along with the visible image as input will improve the model marginally. Using the post-processed target in the training stage helps the model to converge to more realistic results.

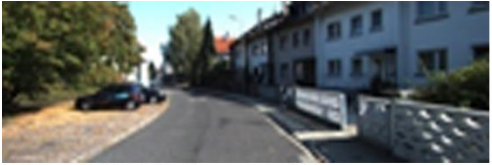
RGB Reference Frame

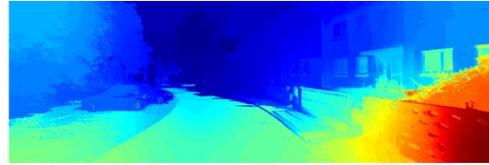
GT Computed by Stereo Matching

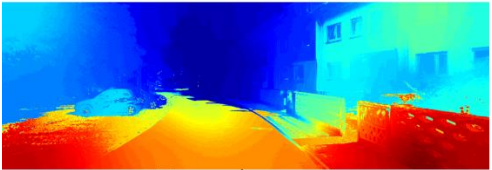
Experiment 1

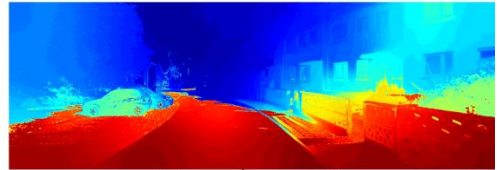
Experiment 2

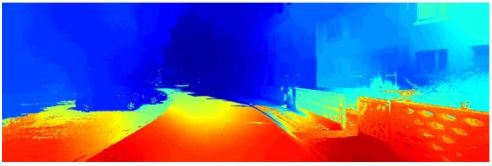
Experiment 3

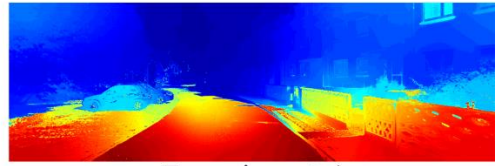
Experiment 4

**Fig. 9** Estimated depth maps from the trained models – example 1

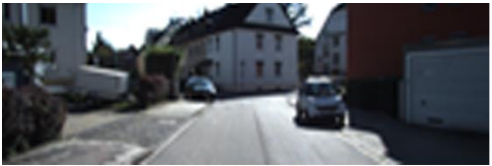
RGB Reference Frame

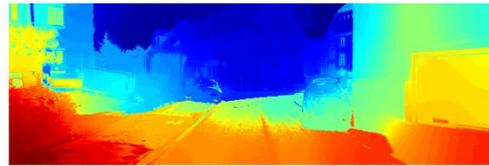
GT Computed by Stereo Matching

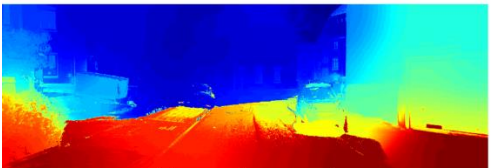
Experiment 1

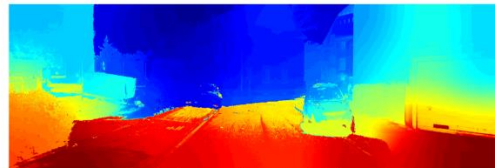
Experiment 2

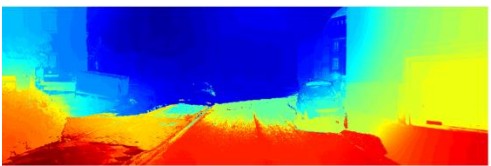
Experiment 3

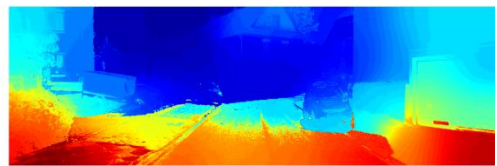
Experiment 4

**Fig. 10** Estimated depth maps from the trained models – example 2

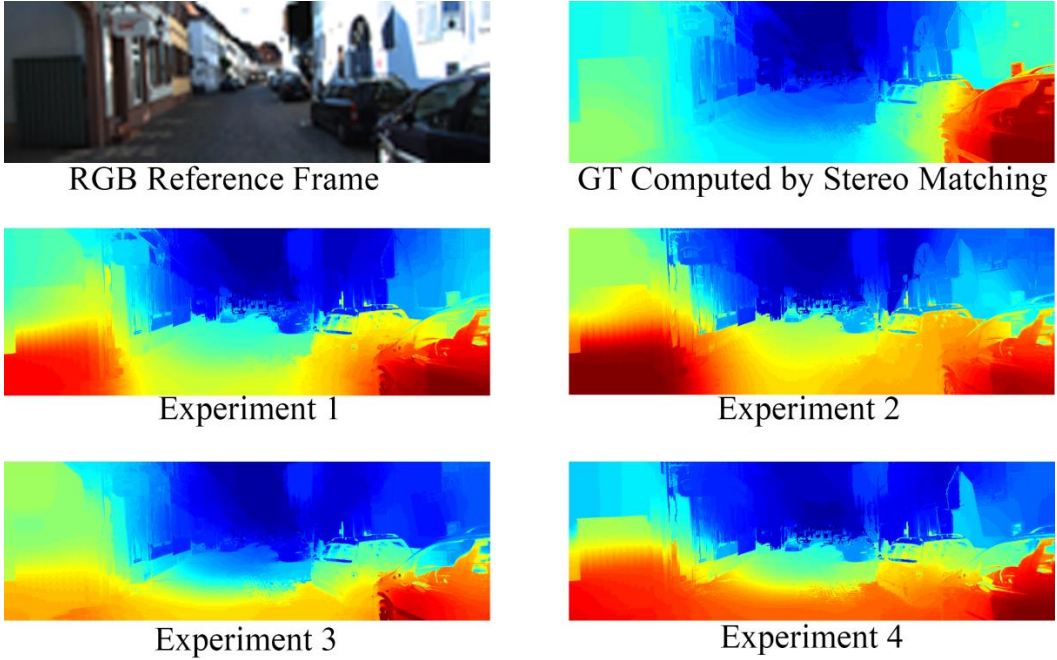

**Fig. 11** Estimated depth maps from the trained models – example 3

Figs. 9-11 indicate that the trained models in this paper are able to estimate depth maps comparable to state-of-the-art stereo matching with structural accuracy and precise depth levels. This is also a result of using the semantic segmentation data and injecting the structural information into the network.

*3.3 Comparing Mono Camera Results with Stereo Matching*

In this section, the results from the mono camera depth estimation given by the proposed method are compared with one of the top-ranked stereo matching methods given in [29]. The ground truth for this comparison is the set of depth maps provided by the KITTI benchmark.

The test images have been forward propagated through the models trained in Sec 2.3 and the best results are compared with the stereo matching technique. The results are shown in Table 4.

The results indicate that using mono camera images and deep learning techniques can provide results which are comparable to stereo matching techniques. As shown in Table 4, the mono camera DNN method was able to provide depth maps similar to the stereo matching methods, represented by PSNR, MSE, MAE, RMSE, and SNR.

Table 4 Numerical comparison between stereo matching and the proposed mono camera model

|      | Stereo Matching [29] | Mono Camera DNN |
|------|----------------------|-----------------|
| PSNR | 14.8234              | 14.3424         |
| MSE  | 0.0351               | 0.0382          |
| RMSE | 0.1845               | 0.1937          |
| SNR  | 4.8836               | 4.4026          |
| MAE  | 0.1017               | 0.1107          |
| SSIM | 0.9966               | 0.9959          |
| UQI  | 0.9353               | 0.9234          |
| PCC  | 0.823                | 0.7687          |

Having close values for SSIM (0.9966 and 0.9959 in the range [0,1]) and UQI (0.9353 and 0.9234 in the range [0,1]) shows how the mono camera DNN method is able to preserve the structural information, as compared to the Stereo Matching method.

## 3.4 Comparison against Other Monocular Depth Estimation Methods

In this section, the proposed network is compared again the method presented in [24,26-28]. Table 5 represents the performance of the proposed network compared to the state of the art methods based on seven metrics including Absolute Relative difference, Squared Relative difference, and RMSE/RMSE log. These numbers indicate that the unsupervised CNN proposed by Godard et al. [28] outperforms the others because of the left-right disparity consistency term which allows the network to optimize the disparity values based on both left and right images. However, we believe that the proposed network has a competitive performance compared to the studied methods considering the fact that our models are trained using only left image without taking into account the influence of the right disparity values.

Table 5 Results on the KITTI 2015 stereo 200 training set disparity images.

| Method | Supervised | Dataset | Abs Rel | Sq Rel | RMSE | RMSE log | $\delta < 1.25$ | $\delta < 1.25^2$ | $\delta < 1.25^3$ |
|---|---|---|---|---|---|---|---|---|---|
| Eigen et al. [26] Coarse | Yes | KITTI | 0.361 | 4.826 | 8.102 | 0.377 | 0.638 | 0.804 | 0.894 |
| Eigen et al. [26] Fine | Yes | KITTI | 0.203 | 1.548 | 6.307 | 0.282 | 0.702 | 0.890 | 0.958 |
| Liu et al. [24] DCNF-FCSP FT | Yes | KITTI | 0.201 | 1.584 | 6.471 | 0.273 | 0.68 | 0.898 | **0.967** |
| Garg et al. [27] L12 Aug 8× cap 50m | No | KITTI | 0.169 | 1.080 | 5.104 | 0.273 | 0.740 | 0.904 | 0.962 |
| Godard et al. [28] | No | KITTI | **0.148** | 1.344 | 5.927 | **0.247** | **0.803** | **0.922** | 0.964 |
| Ours | Yes | KITTI | 0.288 | **1.065** | **4.071** | 0.401 | 0.51 | 0.77 | 0.893 |

Lower is better    Higher is better

## 3.5 Comparing Running Times

In this section, the computational time of the proposed method is compared against the stereo matching methods provided in Table 1. The evaluations indicate that the proposed method is able to perform at a rate of ~1.23 sec/MP on a desktop computer equipped with i7 2600 CPU @ 3.4 GHz and 16GB of RAM.

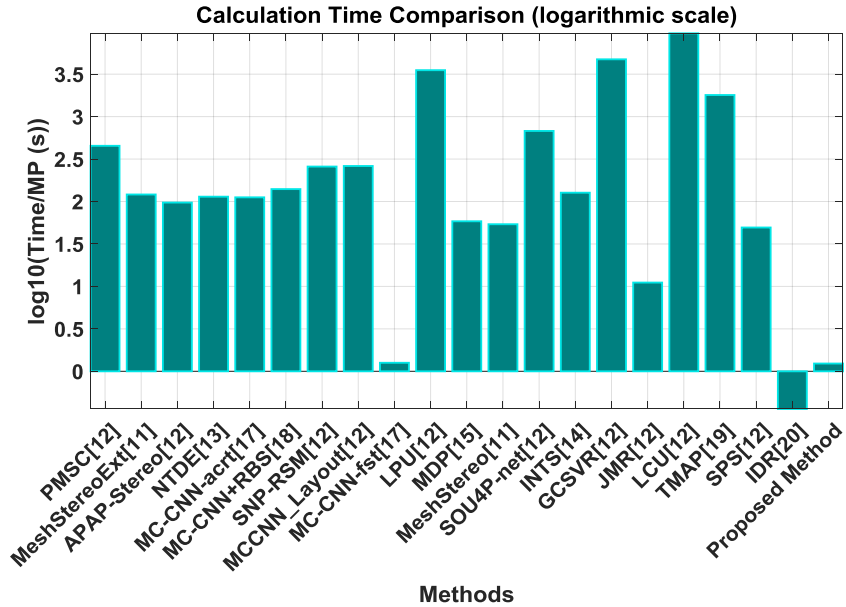

**Fig. 12** Comparison of computational time in logarithmic scale

Fig. 12 shows the comparison of the computational times. The comparison is done in a logarithmic scale due to the large range of computational times between different methods.

## 4 Conclusion and Discussion

In this paper, we have introduced the use of the *Semi Parallel Deep Neural Networks* (SPDNN) method. An SPDNN is a network topology developed using a graph theory optimization of a set of independently optimized CNNs, each targeted at a specific aspect of the more general classification problem. For depth estimation from a monocular set up, a model including fully-connected topology optimized for fine features is combined with a series of max-pooled topologies. The optimized SPDNN topology is re-trained on the full training dataset and converges to an improved set of network weights. Here we used this design strategy to train an accurate model for estimating depth from monocular images.

In this work, 8 different deep neural networks have been mixed and merged using the SPDNN method in order to take advantage of each network's qualities. The mixed network architecture was then trained in four separate scenarios wherein each scenario uses a different set of inputs and targets during training. Four distinct models have been trained. The pixel-wise segmentation and depth estimations given in [29] were used to provide samples for use in the training stage. The KITTI benchmark was used for training and experimental purposes.

Each model was evaluated in two sections, first against the ground truth provided by the benchmark, and secondly against the disparity maps computed by the stereo matching method (Sec 3.1 and 3.2). The results show that using the post-processed depth map presented in [29] for training the network results in more precise models and adding the semantic segmentation of the input frame to the input helps the network preserve the structural information in the output depth map. The results in Sec 3.2 show how close the proposed depth estimation using mono camera can be to the stereo matching method. The semantic segmentation information helps the network converge to the stereo matching results, although the improvement is marginal in this case. The results of the third comparisons in Sec 3.3 show a slightly higher accuracy obtained by employing the stereo matching technique, but our results demonstrate that there is not a big difference between the depths from the models trained by proposed DNN and the values computed by stereo matching. The numerical results of this evaluation show the similarity between the mono camera using DNN method and the stereo matching method, and also the power of the presented method in preserving the structural information in the output depth map.

An important advantage of these models is the processing time of ~1.23 sec/MP. This is equal to 38 fps for an input image of size (80×264) on an i7 2600 CPU @ 3.4 GHz and 16GB of RAM. This makes the model suitable for providing depth estimation in real time. This performance is comparable to the stereo methods MC-CNN-fst [17] and JMR [12], which are 37 fps and 4 fps respectively for the same size of the image, taking advantage of GPU computation power (NVIDIA GTX TITAN X and GTX 980 respectively). The IDR method [20] can give up to 131 fps for the same image size by using an NVIDIA GeForce TITAN Black GPU and CUDA C++ implementation, but the performance on CPU is not given by the authors, so any comparisons with this method would be unfair.

Using pixel-wise segmentation as one of the inputs of the network slightly increased the accuracy of the models, and also helped the model preserve the structural details of the input image. However, it also brought some artifacts, such as wrong depth patches on the surfaces. The evaluation results also illustrate the higher accuracy of the models where a post-processed depth map was used as the target in the training procedure.

*4.1 Future Works and Improvements*

The model presented in this work is still a big model to implement in low power consumer electronic devices (e.g., handheld devices). Future work will include a smaller design which is able to perform as well as the presented model. The other consideration for the current method is the training data size (which is always the biggest consideration with deep learning approaches). The amount of stereo data available in the databases is usually not big enough to train a deep neural network. The augmentation techniques can help to expand databases, but the amount of extra information they provide is limited. Providing a larger set with accurate depth maps will improve the results significantly.

The SPDNN approach is currently being to other problems and is giving promising results on both classification and regression problems. Those results will be presented in future publications.

*Acknowledgments*

The research work presented here was funded under the Strategic Partnership Program of Science Foundation Ireland (SFI) and co-funded by SFI and FotoNation Ltd. Project ID: 13/SPP/I2868 on "*Next Generation Imaging for Smartphone and Embedded Platforms*". This work is also supported by an Irish Research Council Employment Based Programme Award. Project ID: EBPPG/2016/280.

**Appendix A: Network Design**

*A.1: Individual Networks for Depth Analysis*

The network shown in Fig. 13 is a deep fully convolutional neural network (A fully convolutional neural network is a network wherein all the layers are convolutional layers) with no pooling and no padding. Therefore, no information loss occurs inside the network, as there is no bottleneck or data compression; this network is able to preserve the details of the input samples. But the main problem is that this model is unable to find big objects and coarse features in the image. In order to solve this problem, three other networks have been designed as shown in Figs. 14-16. These three networks take advantage of the max-pooling layers to gain transition invariance and also to recognize bigger objects and coarser features inside the image. These networks use 2×2, 4×4, and 8×8 max-pooling operators, respectively. Larger pooling kernels allow coarser features to be detected by the network. The main problem with these networks was that the spatial details vanished as a result of data compression in pooling layers.

After several attempts of designing different networks, the observations showed that in order to estimate the depth from an image, the network needed to see the whole image as one object. To do that it requires the kernel to be the same size as the image in at least one layer that is equivalent to a fully connected layer inside the network.

In fully connected layers each neuron is connected to all neurons in the previous/next layer. Due to the computationally prohibitive nature of training fully connected layers, and their tendency to cause overfitting, it is desirable to reduce the number of these connections. Adding fully connected layers results in a very tight bottleneck, which seems to be crucial for the depth estimation task, but also causes the majority of the details in the image to be lost. In Figs. 17-20 the networks with fully connected layers are shown. These networks correspond to networks in Figs. 13-16 but with convolutional layers replaced with fully connected layers on the right-hand side of the network. Using different pooling sizes before the fully connected layer will cause the network to extract different levels of features, but all these configurations introduce loss of detail.

Each of these eight configurations has its own advantages and shortcomings, from missing the coarse features to missing the details. None of these designs converged to a reasonable depth estimation model.

The main idea of the SPDNN method is to mix and merge these networks and generate a single model which includes all the layers of the original models in order to be able to preserve the details and also detect the bigger objects in the scene for the depth estimation task.

*A.2: The SPDNN Parallelization Methodology*

*A.2.1: Graph Contraction*

A consideration while parallelizing neural networks is that having the same structure of layers with the same distance from the input, might lead all the layers to converge to similar values. For example, the first layer in all of the networks shown in Figs. 13-20 is a 2D convolutional layer with a 3×3 kernel.

The SPDNN idea uses graph contraction to merge several neural networks. The first step is to turn each network into a graph in which it is necessary to consider each layer of the network as a node in the graph. Each graph starts with the input node and ends with output node. The nodes in the graph are connected based on the connections in the corresponding layer of the network. Note that the pooling and un-pooling layers are not represented as nodes in the graph, but their properties will stay with the graph labels, which will be explained later.

Figs. 13-20 presents the networks and their corresponding compressed graphs. Two properties are assigned to each node in the graph. The first property is the layer structure, and the second one is the distance of the current node to the input node. To convert the network into a graph, a labeling scheme is required. The proposed labelling scheme uses different signs for different layer structures, C for convolutional layer (for example 3C mean a convolutional layer with 3×3 kernel), F for fully connected layer (for example 30F means a fully connected layer with 30 neurons) and P for pooling property (for example 4P means that the data has been pooled by the factor of 4 in this layer).

Some properties, like convolutional and fully connected layers, occur in a specific node, but pooling and un-pooling operations will stick with the data to the next layers. The pooling property stays with the data except when an un-pooling or a fully connected layer is reached. For example, a node with the label (3C8P, 4) corresponds to a convolutional layer with a 3×3 kernel, the 8P portion of this label indicates that the data has undergone 8×8 pooling and the 4 at the end indicates that this label is at a distance of 4 from the input layer. The corresponding graphs, with assigned labels for each network, are illustrated in Figs. 13-20.

The next step is to put all these graphs in a parallel format sharing a single input and single output node. Fig. 21 shows the graph in this step.

In order to merge layers with the same structure and the same distance from the input node, nodes with the exact same properties are labeled with the same letters. For example, all the nodes with properties (3C, 1) are labeled with letter A, and all the nodes with the properties (3C2P, 4) are labeled K, and so on.

The next step is to apply graph contraction on the parallelized graph. In the graph contraction procedure, the nodes with the same label are merged to a single node while saving their connections to the previous/next nodes. For instance, all the nodes with label A are merged into one node, but its connection to the input node and also nodes B, C, D, and E are preserved. The contracted version of the graph in Fig. 21 is shown in Fig. 22.

Afterwards, the graph has to be converted back to the neural network structure. In order to do this process, the preserved structural properties of each node are used. For example node C is a 3×3 convolutional layer which has experienced a pooling operation. Note that the pooling quality will be recalled from the original network.

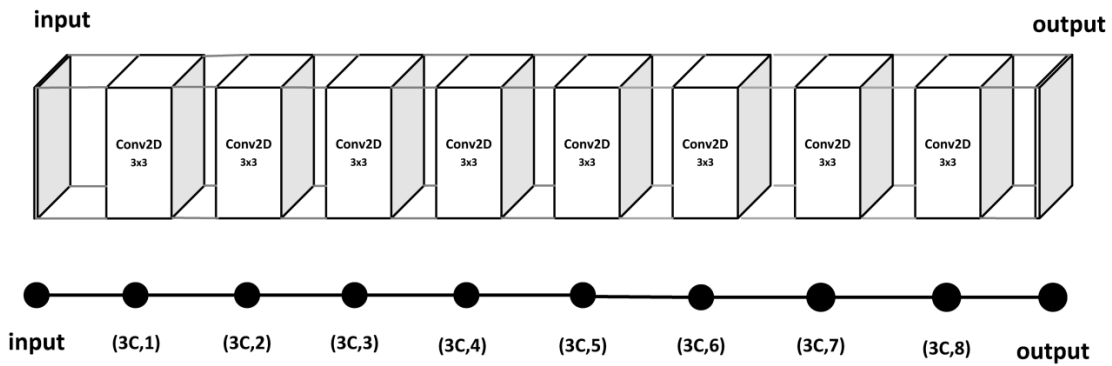

**Fig. 13** Top row: network 1, Bottom row: graph corresponds to network1.

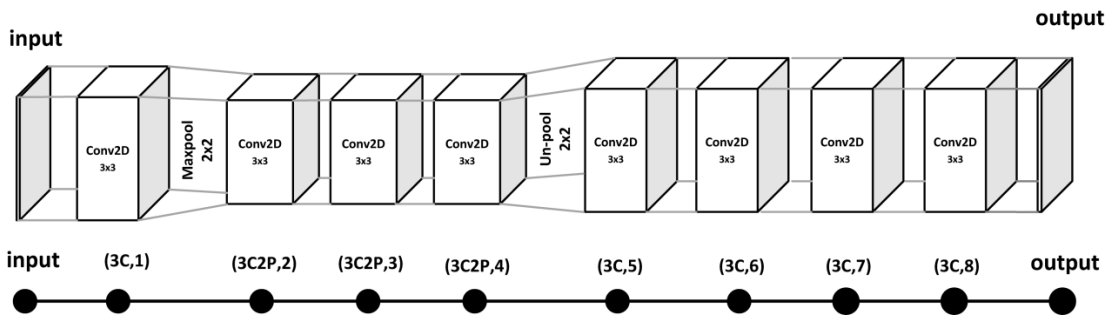

**Fig. 14** Top row: network 2, Bottom row: graph corresponds to network2.

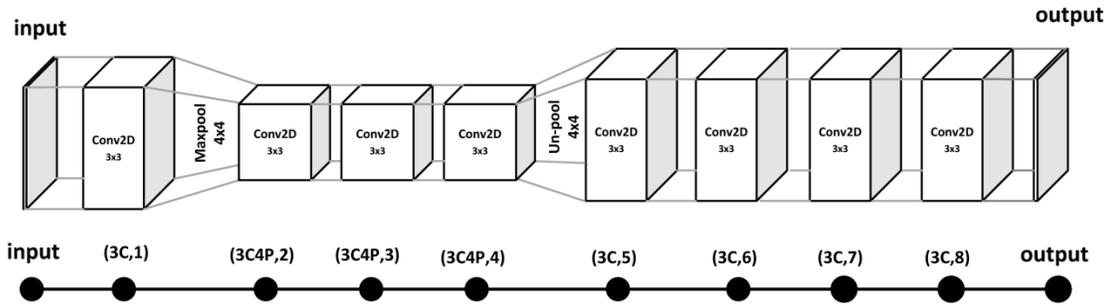

**Fig. 15** Top row: network 3, Bottom row: graph corresponds to network3.

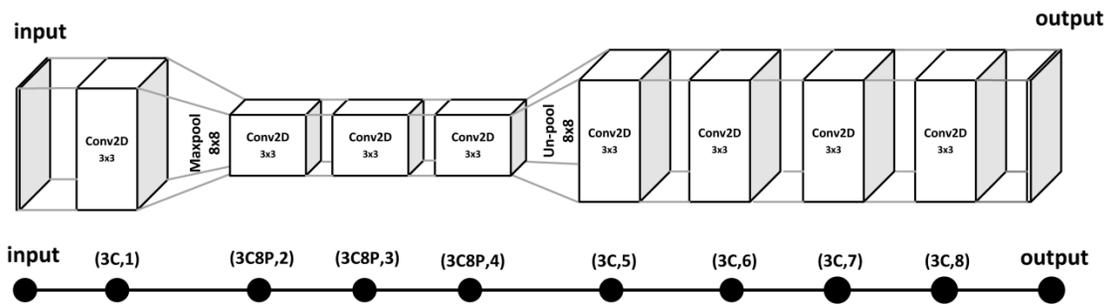

**Fig. 16** Top row: network 4, Bottom row: graph corresponds to network4.

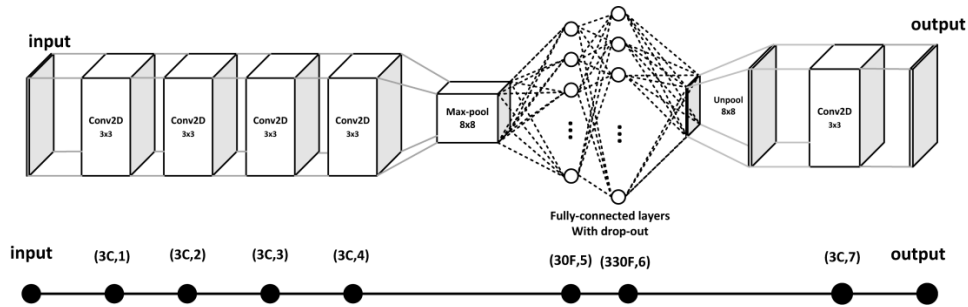

**Fig. 17** Top row: network 5, Bottom row: graph corresponds to network5.

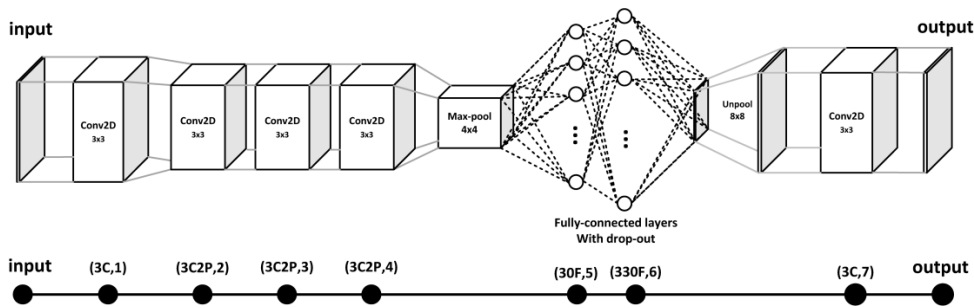

**Fig. 18** Top row: network 5, Bottom row: graph corresponds to network6.

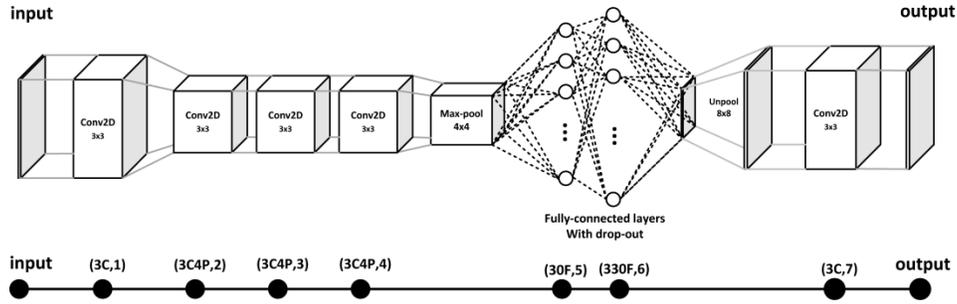

**Fig. 19** Top row: network 7, Bottom row: graph corresponds to network7.

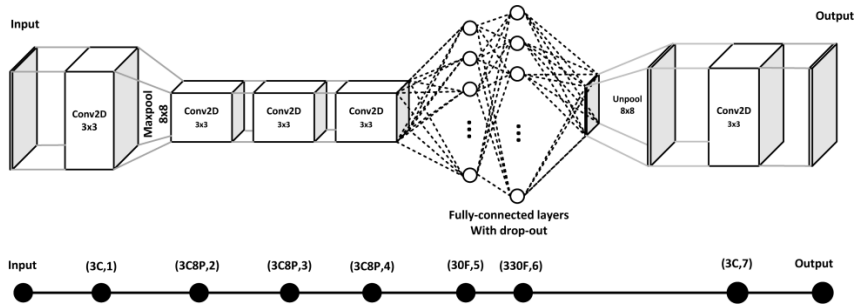

**Fig. 20** Top row: network 8, Bottom row: graph corresponds to network8.

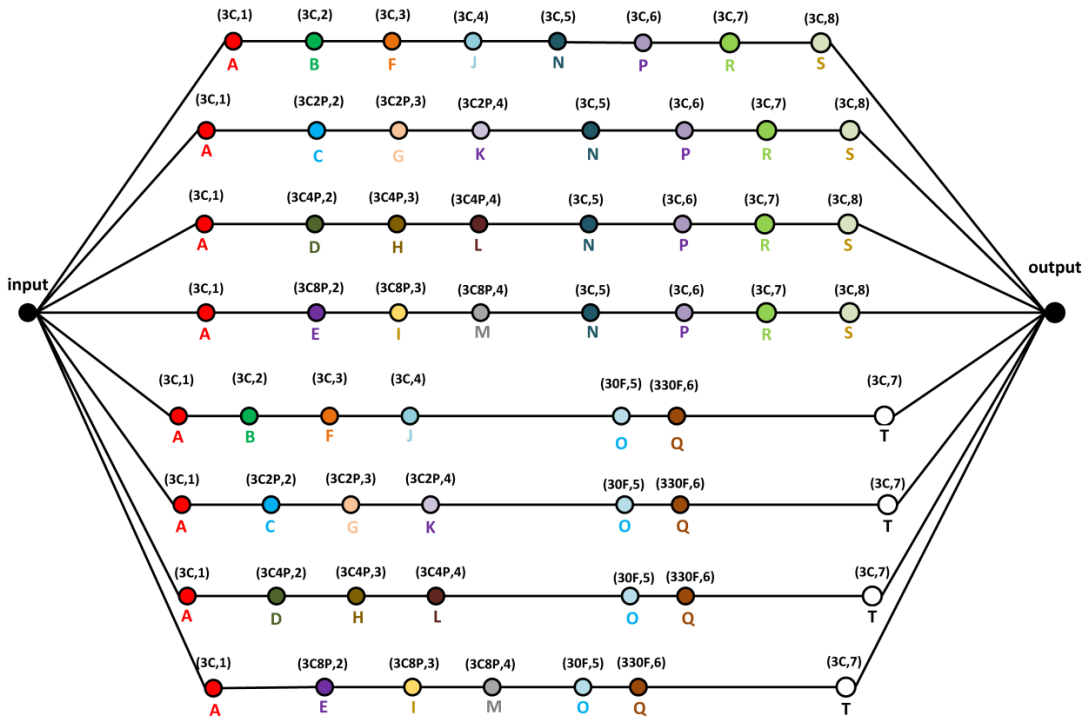

**Fig. 21** Parallelized version of the graphs shown in Figs. 13-20 sharing a single input node and single output node

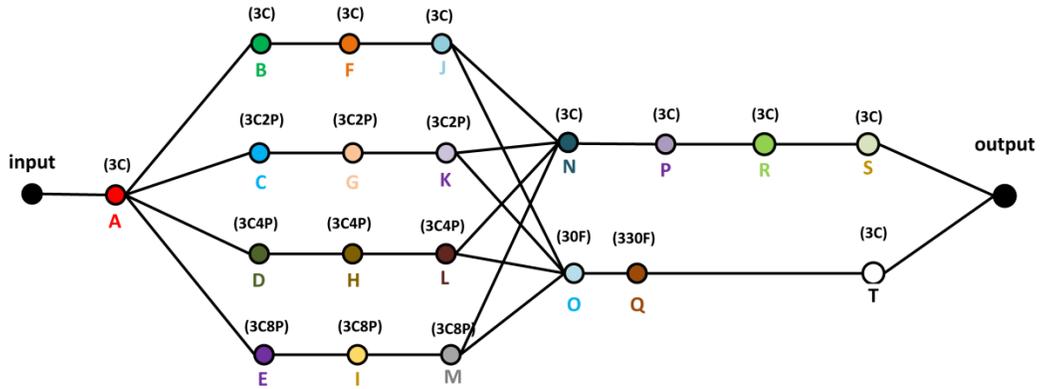

**Fig. 22** Contracted version of the big graph shown in Fig. 21

The concatenation layer is used in the neural network in order to implement the nodes wherein several other nodes lead to one node. For example, in nodes N and O the outputs of nodes J, K, L, and M are concatenated with the pooling qualities taken from their original networks.

The graph is translated back to a deep neural network. The network correspond to the graph in Fig. 22 is shown in Fig. 2.

*A.3: SPDNN: How it Works and why it is Effective?*

One might ask why the SPDNN approach is effective and what the difference is between this approach and other mixing approaches. Here the model designed by the SPDNN scheme is investigated in the forward and back propagation steps. The key component is in the back-propagation step where the parameters in parallel layers influence each other. These two steps are described below:

Forward propagation: Consider the network designed by the SPDNN approach shown in Fig. 23. This exemplary network is made of five sub-networks. Just the general view of the network is shown in this figure and the layers' details are ignored since the main goal is to show the information flow within the whole network.

When the input samples are fed into the network, the data travels through the network along three different paths shown in Fig. 24.

At this stage the parallel networks are blind to each other, i.e., the networks placed in parallel do not share any information with each other. As shown in Fig. 24 the data traveling in Sub-Net 1 and Sub-Net 2 are not influenced by each other since they do not share any path together, as in Sub-Net 3 and Sub-Net 4.

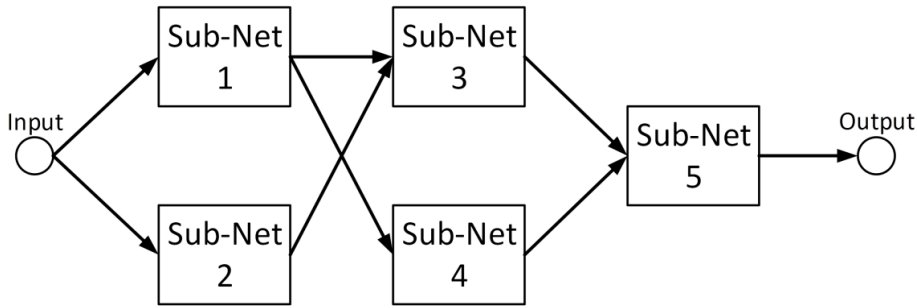

**Fig. 23** A network designed using the SPDNN approach. It contains 5 sub-networks placed in parallel and semi-parallel form.

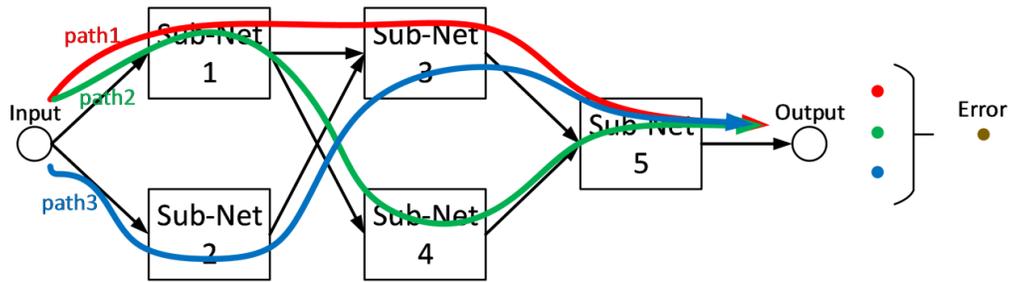

**Fig. 24** Forward propagation inside the SPDNN. There are three different paths on which the information can flow inside the network

Backpropagation: while training the network, the loss function calculated based on the error value at the output of the neural network is a mixed and merged function of the error value corresponding to every data path in the network. In the backpropagation step the parameters inside the network update based on this mixed loss values. i.e., this value back-propagates throughout the whole network as it is shown in Fig. 25. Therefore, at this stage of training, each subnetwork is influenced by the error value from every data path shown in Fig. 25. This illustrates the way each subnetwork is trained to reduce the error of its own path and also the error from the mixture of all paths.

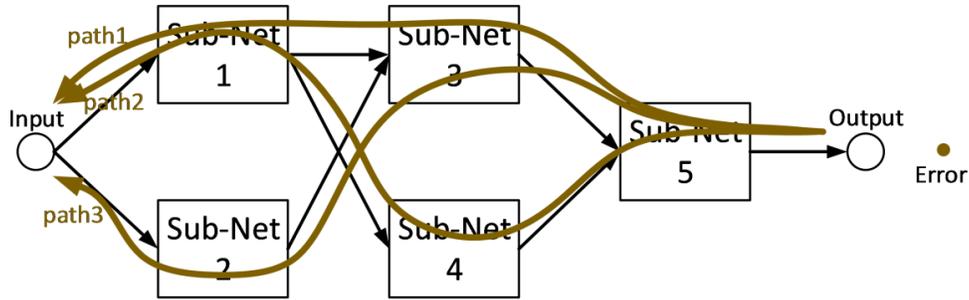

**Fig. 25** Backpropagation for SPDNN. The mixed error is back propagated throughout the network while updating parameters.

The main difference between the SPDNN approach and other mixing approaches, like the voting approach, lies in the back propagation step where different sub-nets are influenced by the error of each other and try to compensate for each other's shortcomings by reducing the final mixed error value. In the voting approach, different classifiers are trained independently of each other and they do not communicate to reduce their total error value.

*A.3.1: SPDNN vs. Inception*

One of the approaches that has superficial similarities to SPDNN is the Inception technique [41]. For clarity, and to aid the reader in understanding, the authors list four significant points of difference between SPDNN and Inception with regard to mixing networks.

1. The main idea in SPDNN is to maintain the overall structure of the networks, but to mix them in a reasonable way. For example, if there is a big kernel such as 13×13 in one of the configurations, the SPDNN method always preserves the structure (13×13 kernel) inside the final network. This contrasts with inception [41], which reduces larger kernels into smaller ones.

2. In the inception method, all the layers are merged into one final layer, which does not happen with the SPDNN approach.

3. The number of the layers in the SPDNN architecture is less than or equal to the number of the layers in the original networks. In contrast, the inception idea aims to increase the number of layers in the network by (it breaks down each layer into several layers with smaller kernels).

The SPDNN idea is to design a new network from existing networks that perform well at some task or subtask while the idea in inception is to design a network from scratch.

## Appendix B: SegNet

SegNet is fully convolutional semantic image segmentation framework presented in [31,32]. This model uses the convolutional layers of the VGG16 network as the encoder of the network and eliminates the fully connected layers, thus reducing the number of trainable parameters from 134M to 14.7M, which represents a reduction of 90% in the number of parameters to be trained. The encoder portion of SegNet consists of 13 convolutional layers with ReLU nonlinearity followed by max-pooling (2×2 window) and stride 2 in order to implement a non-overlapping sliding window. This consecutive max-pooling and striding results in a network configuration that is highly robust to translation in the input image but, has the drawback of losing spatial resolution of the data.

This loss of spatial resolution is not beneficial in segmentation tasks where it is necessary to preserve the boundaries of the input image in the segmented output. To overcome this problem, the following solution is given in [31]. As most of the spatial resolution information is lost in the max-pooling operation, saving the information of the max-pooling indices and using this information in the decoder part of the network preserves the high-frequency information.

Note that for each layer in the encoder portion of the network there is a corresponding decoder layer. The idea of SegNet is that wherever max-pooling is applied to the input data, the index of the feature with the maximum value is preserved. Later these indices will be employed to make a sparse feature space before the de-convolution step, applying the un-pooling step in the decoder part. A batch normalization layer [23] is placed after each convolutional layer to avoid overfitting and to promote faster convergence. Decoder filter banks are not tied to corresponding encoder filters and are trained independently in the SegNet architecture.

**Shabab Bazrafkan** received his B.Sc degree from Urmia University, Urmia, Iran in electrical engineering in 2011 and M.Sc degree from Shiraz University of Technology (SuTECH) in telecommunication engineering, Image processing branch in 2013. Currently he is a PhD student at the National University of Ireland Galway (NUIG) and also works with Xperi Ireland. His field of working is Deep Neural Networks and Neural Network design.

**Hossein Javidnia** received his Master's degree in Information Technology Engineering from the University of Guilan, Iran in 2014. He has started his PhD since 2015 in Electrical Engineering at National University of Ireland, Galway. His current research interests include image processing, machine vision and automotive navigation.

**Joseph Lemley** received a B.S. degree in computer science and the Masters degree in computational science from Central Washington University in 2006 and 2016, respectively. He is currently pursuing the Ph.D. with the National University of Ireland Galway. His field of work is machine learning using deep neural networks for tasks related to computer vision. His Ph.D. is funded by FotoNation, Ltd., under the IRCSET Employment Ph.D. Program.

**Peter Corcoran** (F'10) is a Fellow of IEEE, Editor-in-Chief of IEEE Consumer Electronics Magazine and a Professor with a Personal Chair at the College of Engineering & Informatics at NUI Galway. His research interests include biometrics, cryptography, computational imaging and consumer electronics. He is co-author on 300+ technical publications and co-inventor on more than 250 granted US patents. In addition to his academic career, he is also an occasional entrepreneur, industry consultant and compulsive inventor.

**Caption List**

**Fig. 2** The overview of the trained models in this paper. The semantic segmentation is just used in two experiments

**Fig. 2** The model designed for the depth estimation from monocular images.

**Fig. 3** The repeating technique used in un-pooling layers.

**Fig. 4** Train loss for each experiment

**Fig. 5** Validation loss for each experiment

**Fig. 6** Estimated depth maps from the trained models – example 1

**Fig. 7** Estimated depth maps from the trained models – example 2

**Fig. 8** Estimated depth maps from the trained models – example 3

**Fig. 9** Estimated depth maps from the trained models – example 1

**Fig. 10** Estimated depth maps from the trained models – example 2

**Fig. 11** Estimated depth maps from the trained models – example 3

**Fig. 12** Comparison of computational time in logarithmic scale

**Fig. 13** Top row: network 1, Bottom row: graph corresponds to network1.

**Fig. 14** Top row: network 2, Bottom row: graph corresponds to network2.

**Fig. 15** Top row: network 3, Bottom row: graph corresponds to network3.

**Fig. 16** Top row: network 4, Bottom row: graph corresponds to network4.

**Fig. 17** Top row: network 5, Bottom row: graph corresponds to network5.

**Fig. 18** Top row: network 5, Bottom row: graph corresponds to network6.

**Fig. 19** Top row: network 7, Bottom row: graph corresponds to network7.

**Fig. 20** Top row: network 8, Bottom row: graph corresponds to network8.

**Fig. 21** Parallelized version of the graphs shown in Figs. 13-20 sharing a single input node and single output node

**Fig. 22** Contracted version of the big graph shown in Fig. 21

**Fig. 23** A network designed using the SPDNN approach. It contains 5 sub-networks placed in parallel and semi-parallel form

**Fig. 24** Forward propagation inside the SPDNN. There are three different paths on which the information can flow inside the network

**Fig. 25** Backpropagation for SPDNN. The mixed error is back propagated throughout the network while updating parameters.

**Table 1** Comparison of the performance time between the most accurate stereo matching algorithms

**Table 2** Numerical comparison of the models given the benchmark's ground truth

**Table 3** Numerical comparison of the models given the ground truth from stereo matching

**Table 4** Numerical comparison between stereo matching and the proposed mono camera model

**Table 5** Results on the KITTI 2015 stereo 200 training set disparity images